\documentclass[runningheads]{llncs}
\usepackage{graphicx}
\usepackage{adjustbox}
\usepackage{arxiv}

\usepackage{tikz}
\usepackage{comment}
\usepackage{amsmath,amssymb} 
\usepackage{color}

\usepackage{graphicx}
\usepackage{bbm}
\usepackage{bm}
\usepackage{xspace}
\usepackage[utf8]{inputenc} 
\usepackage{url}            
\usepackage{booktabs}       
\usepackage{amsfonts}       
\usepackage{nicefrac}       
\usepackage{microtype}      
\usepackage{subcaption}
\usepackage{algorithm}
\usepackage{algorithmic}
\usepackage{array}
\usepackage{multirow}
\usepackage{enumitem}
\usepackage{makecell}
\usepackage{gensymb}         
\usepackage{mathtools}
\usepackage{dblfloatfix}
\usepackage{pifont}
\usepackage[permil]{overpic}

\usepackage{titletoc,tocloft}

\newcommand{\xmark}{\ding{55}}

\newcommand{\cmark}{\ding{51}}

\newcommand{\bphi}{\bm{\phi}}

\makeatletter
\DeclareRobustCommand\onedot{\futurelet\@let@token\@onedot}
\def\@onedot{\ifx\@let@token.\else.\null\fi\xspace}

\def\eg{\emph{e.g}\onedot} 
\def\ie{\emph{i.e}\onedot}

\def\etal{\emph{et al}\onedot}
\makeatother

\newcommand{\threesixty}{360$\degree$~}
\def\argmin{\mathop{\mathrm{arg}\, \mathrm{min}}\limits}

\def\argmin{\mathop{\mathrm{arg}\, \mathrm{min}}\limits}

\usepackage{xcolor}
\usepackage{ifthen}
\usepackage{textcomp}
\definecolor{red}{rgb}{1,0,0}
\definecolor{slateblue}{rgb}{0.7,0.35,0.9}
\definecolor{green}{rgb}{0,.9,0}
\definecolor{mahogany}{rgb}{0.75, 0.25, 0.0}
\definecolor{purple}{rgb}{0.6, 0, 0.6}
\definecolor{darkpurple}{rgb}{0.3, 0, 0.3}
\definecolor{darkgreen}{rgb}{0, 0.4, 0}
\definecolor{frenchblue}{rgb}{0.0, 0.45, 0.73}
\definecolor{blue}{rgb}{0.1,0.6,1}
\definecolor{goldenrod}{rgb}{0.65, 0.45, 0.03}
\definecolor{gray}{rgb}{0.5,0.5,0.5}
\definecolor{gold}{rgb}{1.0, 0.874, 0}
\definecolor{silver}{rgb}{0.67,0.67,0.67}
\definecolor{brown}{rgb}{0.8, 0.678, 0.4}

\newcommand{\haowen}[1]{\textcolor{darkpurple}{#1}}

\usepackage[accsupp]{axessibility}  

\usepackage{cite}
\usepackage[pagebackref,breaklinks,colorlinks]{hyperref}
\usepackage[capitalize]{cleveref}
\crefname{section}{Sec.}{Secs.}
\Crefname{section}{Section}{Sections}

\begin{document}
\pagestyle{headings}

\title{Self-supervised \threesixty Room Layout Estimation} 

%
\author{Hao-Wen Ting \qquad
Cheng Sun\qquad
Hwann-Tzong Chen\\
\\
National Tsing Hua University, Hsinchu 30013, Taiwan
}
%
%
\date{}
\maketitle

\begin{abstract}
We present the first self-supervised method to train \threesixty room layout estimation models without any labeled data.
Unlike per-pixel dense depth that provides abundant correspondence constraints, layout representation is sparse and topological,  hindering the use of self-supervised reprojection consistency on images.
To address this issue, we propose {\it Differentiable Layout View Rendering}, which can warp a source image to the target camera pose given the estimated layout from the target image.
As each rendered pixel is differentiable with respect to the estimated layout, we can now train the layout estimation model by minimizing reprojection loss.
Besides, we introduce regularization losses to encourage Manhattan alignment, ceiling-floor alignment, cycle consistency, and layout stretch consistency, which further improve our predictions.
Finally, we present the first self-supervised results on ZilloIndoor and MatterportLayout datasets.
Our approach also shows promising solutions in data-scarce scenarios and active learning, which would have an immediate value in the real estate virtual tour software.
Code is available at \href{https://github.com/joshua049/Stereo-360-Layout}{https://github.com/joshua049/Stereo-360-Layout}.
\end{abstract}

\section{Introduction}
Automatic room layout reconstruction from \threesixty images is an in-demand technique for real-world applications like real estate virtual tours.
It also constitutes a crucial component of holistic 3D understanding~\cite{ZhangCCLZBZ21} and extreme indoor SfM~\cite{ShabaniSOFF21}.
To estimate high-quality room layout, recent methods mostly use deep-learning-based models to predict layout boundaries, corners, or wall segments with steady progress on specialized model design, layout output representation, and training objective formulation.
However, deep models are data-hungry; annotating the layout is a demanding task, which makes fully-supervised layout estimators costly to achieve reasonable results.

One affordable alternative is self-supervised learning with image reprojection consistency, where we can train the models by using unlabeled images with visual overlap.
In the past few years, self-supervised learning for monocular dense depth estimation has achieved good results, with more techniques being proposed to reduce the performance gap to the fully-supervised versions.
However, on the task of \threesixty layout estimation, self-supervised learning is still an unexplored area.
To the best of our knowledge, this work is the first approach toward training a deep layout estimator without labeled data.


A critical challenge of the self-supervised room layout model comes from the difference between the representations of dense depth and room layout.
The estimated depth map can easily be resized and compute the reprojected UV coordinate on new camera poses for backward image warping.
Conversely, the typical representation of room layout is inherently sparse and topological.
For instance, one popular method, HorizonNet~\cite{SunHSC19}, only regresses the positions of wall-floor and wall-ceiling boundaries at each image column.
Using these sparse points to train a model with image reprojection loss can easily be stuck in the local optimum.
The sparse points issue also exists in the recent flatten layout depth regression technique~\cite{WangYSCT21}.
Another category of layout representation is 2D probabilistic maps, \eg, 
floor and ceiling segmentation maps~\cite{PintoreAG20,YangWPWSC19} or layout corner and boundary heat maps~\cite{ZouCSH18}.
Unfortunately, it is unclear how to design a differentiable process to compute the geometry coordinates from the probabilistic map-based predictions, so we opt to adopt models that regress the flatten layout coordinates directly in this work.
In sum, due to the characteristics of layout representation, the existing self-supervised techniques for dense depth are not straightforwardly reproducible for room layout estimation.

To overcome the challenge, we present {\it Differentiable Layout View Rendering} that warps images based on the layout estimation in a differentiable manner.
Specifically, given a target image, we first estimate the wall-floor and wall-ceiling boundary positions for all image columns.
We can then compute the 3D positions of all pixels in an image column using the estimated boundary positions at the column as per the assumptions of the horizontal floor, horizontal ceiling, and vertical wall.
These 3D points can then be projected to a source image and bilinear sample the colors.
As the overall process is differentiable with respect to the estimated layout boundaries, we can end-to-end train a model to minimize the photometric errors between the warped source image and the target image.
Although this approach brings self-supervised learning into the \threesixty room layout estimation task, it is still an ill-posed and challenging problem.
To facilitate this task, our idea is to incorporate prior knowledge and self-consistency into our models.
We propose training losses on the estimated layout boundaries to encourage {\it i)} Manhattan alignment, {\it ii)} ceiling-floor alignment, {\it iii)} consistency between the estimated layout from original images and rendered images, and {\it iv)} consistency between the stretched estimated layout and the estimated layout from stretched image.
The effectiveness of all our designs is verified quantitatively, and, for the first time, we accomplish self-supervised training of \threesixty room layout estimation models.

Our contributions are manifold and summarized as follows:
\begin{itemize}[topsep=0pt]
    \item We are the first to train deep \threesixty room layout models without labels.
    \item The proposed {\it Differentiable Layout View Rendering} bridges the gap between the dense representation of depth and the flattened/sparse representation of \threesixty room layout, which makes the image reprojection loss applicable.
    \item The introduced Manhattan alignment, ceiling-floor alignment, cycle consistency, and layout stretch consistency losses are shown to be effective.
    \item The self-supervised pre-trained model is data-efficient, significantly improving results over the standard ImageNet pre-trained.
    \item Our self-supervised model is also helpful in active learning. Training on the actively-queried labeled images leads to much better results than randomly-selected labeled images.
\end{itemize}

\section{Related work}
\paragraph{\threesixty room layout estimation.}
PanoContext~\cite{ZhangSTX14} is one of the pioneering attempts to estimate the room layout from a single \threesixty image, where conventional features like Orientation Map and Geometric Context are employed to score and select the layout hypothesis.
Later, LayoutNet~\cite{ZouCSH18} demonstrates the effectiveness of deep learning in \threesixty room layout estimation.
Since then, deep-learning-based models and techniques have been thriving in this task.
One aspect of categorizing these methods is whether the predictions are based on probabilistic maps or regression models.
LayoutNet~\cite{ZouCSH18} and CFL~\cite{Fernandez-Labrador20} predict the heat maps of layout corners and boundaries.
DuLa-Net~\cite{YangWPWSC19} proposes to use bird's-eye views to sidestep \threesixty distortions and predict the ceiling-floor binary segmentation maps, as the follow-up AtlantaNet~\cite{PintoreAG20} seeks to improve the network architecture.
On the other hand, HorizonNet~\cite{SunHSC19} makes good use of the gravity alignment (with image y-axis) prior and proposes to regress the wall-floor and wall-ceiling boundary positions at each image column.
Their sequel HoHoNet~\cite{SunSC21} devises an advanced architecture and extends to more modalities.
Recently, LED$^2$-Net~\cite{WangYSCT21} computes regression losses using the rendered layout depth instead of the image positions, achieving even better results.
The differentiable depth rendering module in LED$^2$-Net only considers the distance to walls of each image column, which is still sparse and is only used in training losses.
Our {\em differentiable layout view rendering} module computes the global 3D coordinates of all pixels for image reprojection. Due to the limitation of space, we only mention a part of the relevant methods. We refer interested readers to Zou \etal~\cite{ZouSPCSWCH21} for a comprehensive review of the recent progress of the \threesixty room layout estimation models.

One challenge of applying the probabilistic-map-based layout predictions to self-supervised learning is that non-differentiable operations (\eg, peak-finding, binarization, contour-finding~\cite{SuzukiA85}) are typically required as the first step to form layout polygons.
Conversely, regression-based predictions can form polygons by simply connecting the adjacent regressed points.
Thus, we opt to use the regression-based layout representation~\cite{SunHSC19} to devise our differentiable rendering module for its inherently differentiable property and outstanding performance~\cite{WangYSCT21,ZouSPCSWCH21} on complex scenes.

SSLayout360~\cite{Tran21} uses mean-teacher to simultaneously learn from labeled and unlabeled data, showing the effectiveness of semi-supervised learning in \threesixty layout estimation, especially when labeled data is scarce.
Our self-supervised approach improves the predictions by providing better weight-initialization and does not have to change the training pipeline for the unlabeled data.
Combining self-supervised and semi-supervised learning could be helpful~\cite{ChenKSNH20}; however, SSLayout360 does not release training code at the time of this preprint being done, and investigating the combination is out of the scope of this work.

\paragraph{Self-supervised depth estimation.}
As densely labeled depth data are costly to acquire, self-supervised learning has appeared as an affordable alternative to learning from large-scale unlabeled images with visual overlap.
Two popular forms of self-supervised learning for depth estimation are {\it i)} learning from stereo pairs with known relative poses~\cite{GargKC016,GodardAB17,XieGF16} and {\it ii)} learning from monocular videos with estimated poses between frames~\cite{GodardAFB19,WangBZL18,YangWWXN18,YangWXZN18,YinS18,ZhouBSL17}.
By minimizing the image reprojection errors, the models should learn to estimate depth for each pixel.
However, these methods are not directly applicable for \threesixty layout estimation due to the sparse representation of layout.
We address this issue by the proposed {\em Differentiable Layout View Rendering}.
Besides, we consider a more constrained capturing setup~\cite{CruzHLKBK21} where each room is captured by multiple \threesixty views with known relative poses and constant distance to the floor.

\section{Approach}

\begin{figure}[t]
    \centering
    \includegraphics[width=\linewidth]{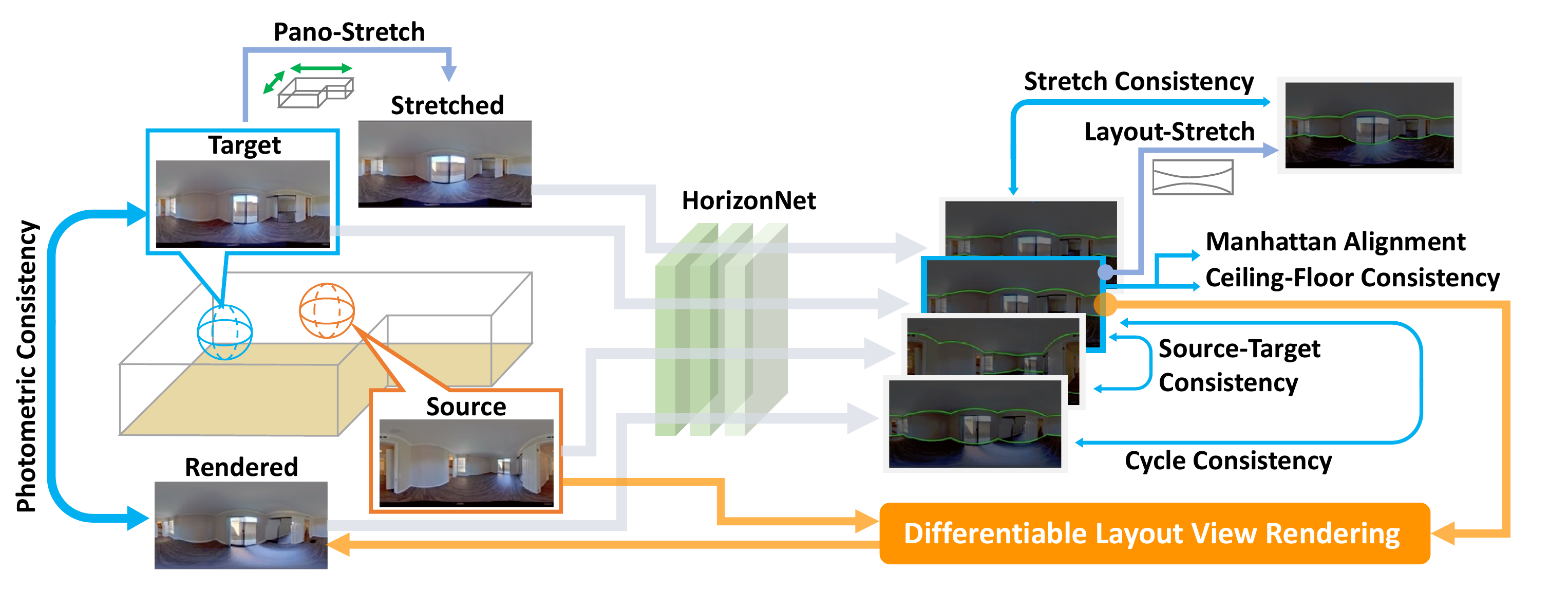}
    \caption{The proposed self-supervised scheme for \threesixty room layout estimation.}
    \label{fig:overview}
\end{figure}

\subsection{Preliminaries}
We demonstrate the proposed self-supervised learning mechanism for \threesixty room layout via HorizonNet~\cite{SunHSC19}.
First, the floor-plane direction alignment~\cite{ZouCSH18} algorithm is applied to the input image as a standard pre-processing step, which facilitates layout inference.
Given an aligned image $\bm{x}\in\mathbb{R}^{H\times W\times 3}$, HorizonNet then regresses the viewing angles $\phi^{\text{(f)}}$ and $ \phi^{\text{(c)}}$ of wall-floor and wall-ceiling boundaries at each image column, where we express them as $W$-dimensional vectors $\bm{\phi}^{\text{(f)}}\in(-\pi/2,0)^{W}, \bm{\phi}^{\text{(c)}}\in(0, \pi/2)^{W}$ for the overall $W$ image columns.
We discard the auxiliary branch for horizontal wall-wall classification as we do not figure out how to train it without labels, while $\bm{\phi}^{\text{(f)}}, \bm{\phi}^{\text{(c)}}$ are enough to represent the room layout as shown in recent work~\cite{PintoreAG20,YangWPWSC19,WangYSCT21}.
In supervised learning, the labeled corners can be used to derive the ground-truth boundaries for supervision.

\subsection{Assumptions} \label{ssec:assumption}
To train a self-supervised \threesixty room layout model, we assume that each room is captured from more than one viewpoints with constant distance to the floor and known relative poses.
We also make the assumptions adopted by most previous arts, \ie, one horizontal floor, one horizontal ceiling, vertical walls, and all \threesixty images are pre-processed by Manhattan alignment algorithm.

\subsection{Differentiable Layout View Rendering} \label{ssec:DLVR}
\paragraph{Overview.}
Our setting of self-supervised learning does not need layout annotations. We only assume that multiple panoramas of a room are available.
The key idea is to train the layout estimation model by maximizing image reprojection consistencies between different viewpoints.
To this end, we present \emph{Differentiable Layout View Rendering}, which warps a source image $\bm{x}^{\text{(s)}}$ to the target viewpoint using the estimated wall-floor $\bm{\phi}^{\text{(f)}}$ and wall-ceiling $\bm{\phi}^{\text{(c)}}$ boundaries from the target image $\bm{x}^{\text{(t)}}$.
Besides, the rendered target image $\bm{x}^{\text{(s}\rightarrow\text{t)}}$ is differentiable with respect to $\bm{\phi}^{\text{(f)}}, \bm{\phi}^{\text{(c)}}$ so it is end-to-end trainable.
Our approach is illustrated in \cref{fig:overview} and detailed below.

\paragraph{Pixel indices to UV coordinates.}
The UV coordinate system in this work is
\begin{equation}
    \bm{u}_{ji} = i / W\cdot 2\pi~,~ \bm{v}_{ji} = (0.5 - j / H)\cdot\pi~,
\end{equation}
where $H, W$ are image height and width, and $j, i$ are row index and column index.

\paragraph{Constrained prediction ranges.}
The original boundary regression of HorizonNet is floating point, and the model can finally learn the proper output range with supervised learning.
However, the unconstrained boundaries could make the training unstable if we adopt self-supervised learning, \eg, rendering a wall behind the viewing direction or too far away.
To address this concern, we apply sigmoid with a hand-crafted scaling to ensure proper ranges for the regressed boundaries.
Specifically, at each image column we obtain 
\begin{equation} \label{eq:output_proc}
    \phi^{\text{(f)}} = -0.5\pi \cdot \operatorname{Sigmoid}\left(\mathring{\phi}^{\text{(f)}}\right),~
    \phi^{\text{(c)}} = 0.5\pi \cdot \operatorname{Sigmoid}\left(\mathring{\phi}^{\text{(c)}}\right)~,
\end{equation}
where $\mathring{\phi}$ denotes HorizonNet's unconstrained output.

\paragraph{Inference of distances to the floor and ceiling.}
The distances from the camera to the horizontal floor plane $z=z^{\text{(f)}}$ and ceiling plane $z=z^{\text{(c)}}$ are necessary to the inference of 3D layout.
In this work, we reconstruct up-to-a-scale 3D and set $z^{\text{(f)}}=-1$ for the entire scene.
In case the camera height $h^{\text{(camera)}}$ and room height $h^{\text{(room)}}$ are provided, we set $z^{\text{(c)}}=\frac{h^{\text{(room)}}-h^{\text{(camera)}}}{h^{\text{(camera)}}}$.
To enable a less constrained data setup when $h^{\text{(camera)}}, h^{\text{(room)}}$ are not annotated, the distance to ceiling is determined by minimizing the mean squared error of the estimator $z^{\text{(f)}} \cot{\phi_i^{\text{(f)}}}  \tan{ \phi_i^{\text{(c)}}} $ derived from $\phi^{\text{(f)}}, \phi^{\text{(c)}}$ at each column $i$.
Specifically, we plug in $z^{\text{(f)}}=-1$ and solve the MSE for $z^{\text{(c)}}$ as
\begin{subequations} \label{eq:zc}
\begin{align}
    z^{\text{(c)}} &= \argmin_{z>0}{ \sum\nolimits_{i=1}^{W}\left(   - \cot{\phi_i^{\text{(f)}}}  \tan{ \phi_i^{\text{(c)}}}  - z \right)^2 }  \label{eq:zc_opt} \\
    &=  \frac{1}{W} \sum\nolimits_{i=1}^{W}  -\cot{\phi_i^{\text{(f)}}}   \tan{ \phi_i^{\text{(c)}}}   ~, \label{eq:zc_ans}
\end{align}
\end{subequations}
where the \cref{eq:zc_ans} is the closed form solution of the optimization problem in \cref{eq:zc_opt}.
See the appendix for an illustration and detail explanation.

\begin{figure}
    \centering
    \begin{subfigure}[b]{0.55\textwidth}
    \centering
    \includegraphics[width=.8\linewidth]{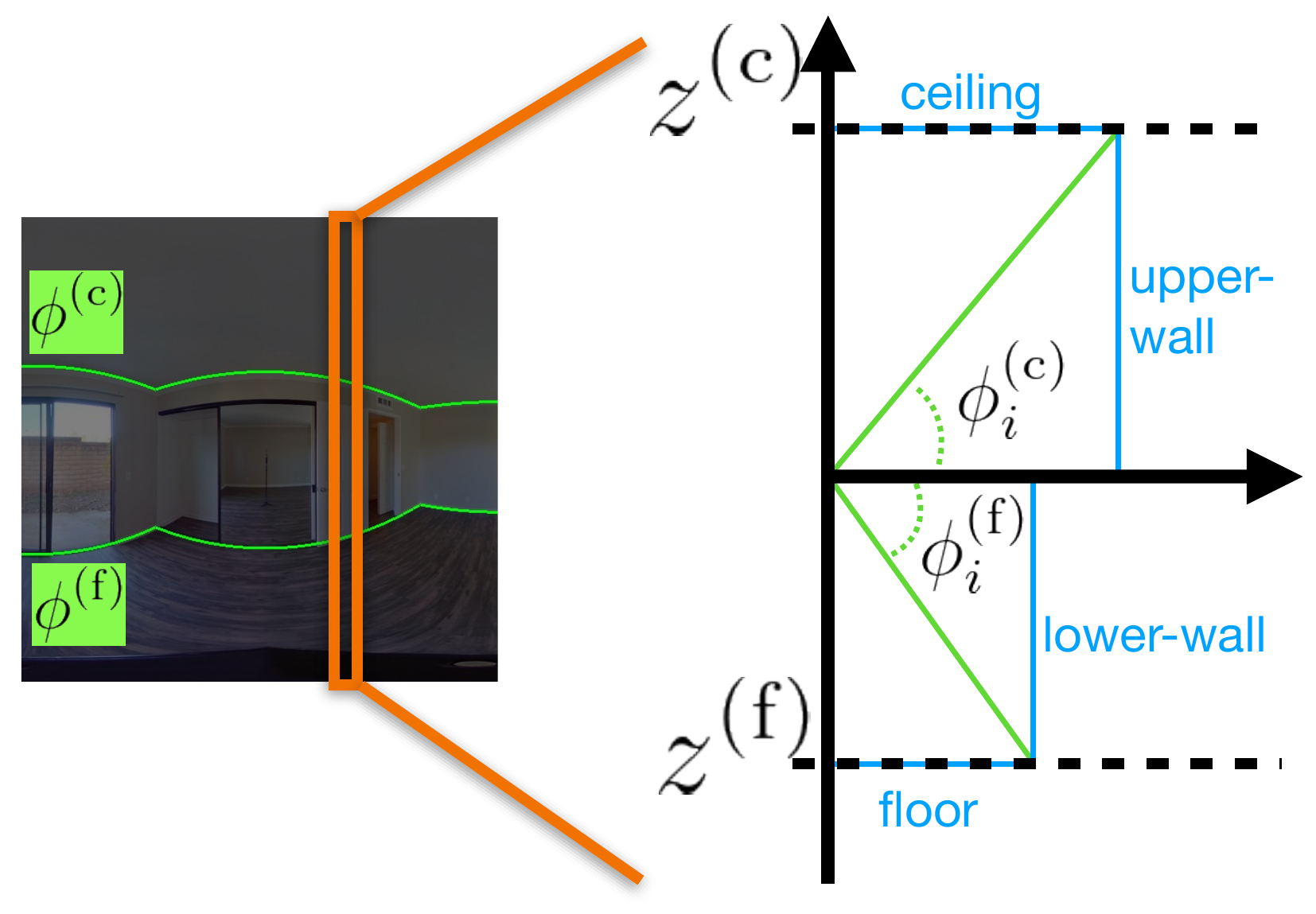}
    \caption{Illustration of the rendered layout (\textcolor{blue}{blue solid lines} for ceiling, upper-wall, lower-wall, and floor) of an image column by our differentiable rendering module (\cref{ssec:DLVR}).}
    \label{fig:sphere}
    \end{subfigure}
    \hfill
    \begin{subfigure}[b]{0.4\textwidth}
    \centering
    \includegraphics[width=\linewidth]{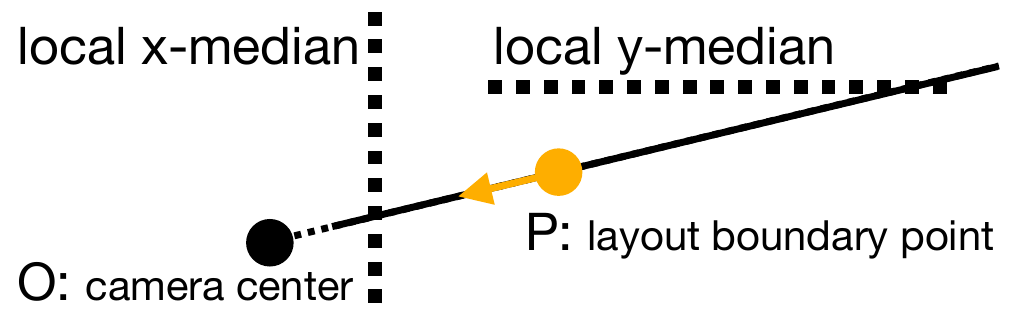}
    \caption{Our loss encourages Manhattan alignment by pulling the regressed layout boundary point $P$ toward the closest local x- or y-median. As $P$ can only move along $\overline{OP}$, the distance is measured by the moving distance on $\overline{OP}$, so, in this example, $P$ is guided to align with x-median.}
    \label{fig:manhattan}
    \end{subfigure}
    \caption{Layout rendering and Manhattan alignment.}
\end{figure}

\paragraph{Inference of 3D coordinates of all pixels as per the estimated layout.}
We parameterize the coordinate $\bm{p}_{ji}$ of a 3D point observed by pixel $(j, i)$ as
\begin{equation} \label{eq:DLVR_3d}
    \bm{p}_{ji} = \bm{d}_{ji} \cdot \left[\cos{\bm{u}_{ji}},~ \sin{\bm{u}_{ji}},~ \tan{\bm{v}_{ji}}\right] ~,
\end{equation}
where $\bm{d}$ is the distance between camera and the observed 3D point projected onto $z=0$.
The value of $\bm{d}_{ji}$ depends on the estimated wall-floor $\phi_i^{\text{(f)}}$ and wall-ceiling $\phi_i^{\text{(c)}}$ boundaries of the same image column $i$.
See \cref{fig:sphere} for a better illustration of the layout rendering.
There are four segments in an image column separated by $0$ and the two boundary points.
The $\bm{d}_{ji}$ of floor pixels, lower-wall pixels, upper-wall pixels, and ceiling pixels can be decided by 
\begin{equation} \label{eq:DLVR_d}
  \bm{d}_{ji} =
    \begin{cases}
    z^{\text{(c)}}\cdot\cot{\bm{v}_{ji}} \:,
    & \bm{v}_{ji}\in [\phi_i^{\text{(c)}},~ 0.5\pi) \:;\\
    z^{\text{(c)}}\cdot\cot{\phi_i^{\text{(c)}}} \:,
    & \bm{v}_{ji}\in (0,~ \phi_i^{\text{(c)}}) \:; \\
    z^{\text{(f)}}\cdot\cot{\phi_i^{\text{(f)}}} \:,
    & \bm{v}_{ji}\in (\phi_i^{\text{(f)}},~ 0) \:; \\
    z^{\text{(f)}}\cdot\cot{\bm{v}_{ji}} \:,
    & \bm{v}_{ji}\in (-0.5\pi,~ \phi_i^{\text{(f)}}] \:.\\
    \end{cases}
\end{equation}
We now get the 3D coordinates of all pixels projected onto the estimated layout by \cref{eq:DLVR_3d} and \cref{eq:DLVR_d}.

As $z^{\text{(f)}}$ is a constant in this work, the gradients from all the floor pixels cannot backpropagate through the network.
On the other hand, ceiling pixels do not have this problem if we use \cref{eq:zc} to determine $z^{\text{(c)}}$.
Therefore, we may alternatively fix $z^{\text{(c)}}$ and solve for $z^{\text{(f)}}$ like \cref{eq:zc} to render another 3D layouts, and then use the two rendered layouts to compute our self-supervised losses.
However, we find in our pilot experiments that using the annotated $z^{\text{(c)}}$ and fixed $z^{\text{(f)}}$ with both ceiling and floor pixels detached from network training yields better results, which suggests that wall pixels are enough to guide network training.
Thus, we choose to keep $z^{\text{(f)}}$ and $z^{\text{(c)}}$ as constant, and include only wall pixels in the update of our model. For completeness' sake, we still render all pixels but leave for future work to continue investigating on this respect.

\paragraph{Warping source images to the target viewpoint.}
After obtaining the 3D coordinates $\bm{p}\in\mathbb{R}^{H\times W\times 3}$ by projecting all pixels from the target image onto the estimated target layout, we can warp a source image $\bm{x}^{\text{(s)}}$ to the target viewpoint using backward image warping:
\begin{equation}
\label{eq:warping}
    \bm{x}_{ji}^{\text{(s}\rightarrow\text{t)}} = \bm{x}^{\text{(s)}}\left\langle\operatorname{EquProj}\left( 
        T_{\text{t}\rightarrow\text{s}} \begin{bmatrix}\bm{p}_{ji}^{\intercal}\\1\end{bmatrix}
    \right)\right\rangle ~,
\end{equation}
where $T_{\text{t}\rightarrow\text{s}}\in\mathbb{R}^{3\times 4}$, derived from the relative pose, transforms the target camera coordinate to the source camera coordinate. The operator  $\operatorname{EquProj}$ performs equirectangular projection, and $\langle\cdot\rangle$ is bilinear sampling operator.

\subsection{Losses for Self Supervision}

\paragraph{Photometric loss.}
We use the Mean Squared Error (MSE) to measure the photometric dissimilarity between the target image $\bm{x}^{\text{(t)}}$ and the rendered image $\bm{x}^{\text{(s}\rightarrow\text{t)}}$ that is warped from the source image to the target viewpoint by \cref{eq:warping}. We superimpose a validity mask $\bm{m}$ on the image to exclude the regions covered by the camera tripod from the computation of the photometric loss:
\begin{equation}
    \mathcal{L}_{\text{photo}} = \frac{1}{N} \sum\nolimits_{k=1}^{N} \bm{m}_k \left\| \bm{x}_{k}^{\text{(s}\rightarrow\text{t)}} - \bm{x}_{k}^{\text{(t)}} \right\|^2 ~,
\end{equation}
where index $k$ iterates through all the $N$ pixels, and $\bm{m}$ activates pixels that are not covered by the camera tripod in both the target and the warped image.

\paragraph{Cycle consistency loss.}
As an extra supervision signal on reprojection, the model predicts the rendered image and enforces its consistency with the target image via the cycle consistency loss:
\begin{equation}
    \mathcal{L}_{\text{cycle}} = \frac{1}{W} \sum\nolimits_{i=1}^{W} \sum\nolimits_{\text{r} \in\{\text{f}, \text{c}\}}
        \left( \phi_i^{\text{(s}\rightarrow\text{t)}(\text{r})}  - \operatorname{detach}\left(\phi_i^{(\text{r})}\right) \right)^2 ~,
\end{equation}
where $W$ is image width and $\bm{\phi}^{\text{(s}\rightarrow\text{t)}}$ is HorizonNet prediction for the rendered image $\bm{x}^{\text{(s}\rightarrow\text{t)}}$.
We use $\operatorname{detach}$ to block the gradient flow of backpropagation as we treat $\bm{\phi}$ as a teacher with a more accurate result.

\paragraph{XY-plane projection.}
We first convert the estimated layout boundary $\bm{\phi}$ to world coordinates to compute the geometry-based losses. As all the boundary points on the floor and ceiling in a room share the same constant z component $z^{\text{(f)}}, z^{\text{(c)}}$, we only care about their projected points onto the XY-plane:
\begin{equation}
    \bm{b}_i^{\text{(r)}} =
    \left[x_i^{\text{(r)}},~ y_i^{\text{(r)}}\right] = z^{\text{(r)}} \cdot \left[
        \cot{\phi_i^{\text{(r)}}}\cos{\bm{u}_{i}},~
        \cot{\phi_i^{\text{(r)}}}\sin{\bm{u}_{i}}\right]
        ~ \text{for} ~  \text{r} \in \{\text{f},\text{c}\}~,
\end{equation}
where $i$ is the column index (see \cref{fig:sphere}). All our geometry-based losses introduced below assume Manhattan world and Manhattan alignment (\cref{ssec:assumption}).

\paragraph{Source-target consistency loss.}
Since the source and target views capture the same room, we encourage their layout boundary points to be as close as possible on the world coordinates by
\begin{equation}
    \mathcal{L}_{\text{src-tgt}} = \sum\nolimits_{\text{r} \in\{\text{f}, \text{c}\}} \operatorname{ChamferDist}\left(
        T_{\text{s}\rightarrow\text{t}}^{\text{(XY)}} \cdot \bm{b}^{\text{(s)(r)}},~
        \bm{b}^{\text{(r)}}
    \right) ~,
\end{equation}
where $T_{\text{s}\rightarrow\text{t}}^{\text{(XY)}}\in\mathbb{R}^{2\times 2}$ transforms the projected points $\bm{b}^{\text{(s)(r)}}\in\mathbb{R}^{2\times W}$ in source camera coordinates to target camera coordinates. We use Chamfer distance to assess the distance between two sets of 2D points.


\paragraph{Ceiling-floor consistency loss.}
The wall-ceiling and wall-floor boundaries of the same image column should be projected onto the same point on the XY-plane, so we minimize
\begin{equation}
    \mathcal{L}_{\text{c-f}} = \frac{1}{W} \sum\nolimits_{i=1}^{W}  \left(x_i^{\text{(c)}} - x_i^{\text{(f)}}\right)^2 + \left(y_i^{\text{(c)}} - y_i^{\text{(f)}}\right)^2  ~,
\end{equation}
to encourage the projected points to be close.


\paragraph{Manhattan alignment loss.}
As all the images are justified by Manhattan alignment algorithm,
we can implement Manhattan alignment regularization by minimizing the discrepancy to local x-median or y-median for each projected boundary point.
However, the naive implementation ignores the fact that the projected point can only move away or toward the camera (illustrated in \cref{fig:manhattan}).
We thus compensate the slope factors for the x and y components:
\begin{equation}
    \mathcal{L}_{\text{M}} = \frac{1}{W} \sum_{i=1}^{W} \sum_{\text{r} \in\{\text{f}, \text{c}\}} \min\left\{
        \left|\left(x_i^{\text{(r)}} - \hat{x}_i^{\text{(r)}}\right)\cdot\sin{\bm{u}_{i}}\right|\,,~
        \left|\left(y_i^{\text{(r)}} - \hat{y}_i^{\text{(r)}}\right)\cdot\cos{\bm{u}_{i}}\right|
    \right\} ~,
\end{equation}
where $\hat{x}$ and $\hat{y}$ are the local x-median and y-median.


\paragraph{Stretch consistency loss.}
We perform the Pano Stretch operation~\cite{SunHSC19} to stretch images, and then compute the MSE between the predicted layout from the stretched image and the stretched layout from the original image:
\begin{equation}
    \mathcal{L}_{\text{stretch}} = \frac{1}{W} \sum\nolimits_{i=1}^{W} \sum\nolimits_{\text{r}\in\{\text{f}, \text{c}\}}
        \left( \phi_i^{\text{(S\_img)}(\text{r})}  - \operatorname{S\_layout}\left(\operatorname{detach}\left(\phi_i^{(\text{r})}\right)\right) \right)^2 ~,
\end{equation}
where $\bm{\phi}^{\text{(S\_img)}}$ is HorizonNet prediction for the randomly stretched image, and $\operatorname{S\_layout}$ stretches the layout with the same stretching parameters as the corresponding image.
Similar to cycle consistency loss, we treat $\bm{\phi}$ as a teacher.

\section{Experiments}

\subsection{Implementation Details}
We adopt almost the same network architecture as HorizonNet~\cite{SunHSC19}. Since our self-supervised pretraining only focuses on boundary prediction, we revise the number of output channels from 3 to 2 so that the model predicts the boundary channels only. For evaluation, due to the lack of corner prediction, we directly regard the boundary prediction as a polygon of $W$ vertices and follow the setting in HorizonNet to calculate the evaluation metrics.
For self-supervised pretraining, we use the Adam optimizer to train the network for 20 epochs with batch size 2 and learning rate $10^{-4}$. For supervised fine-tuning, we use the same optimizer and learning rate, but adjust the number of epochs and batch size to 100 and 4.

\subsection{Datasets}
\paragraph{Zillow Indoor Dataset (ZInD)~\cite{CruzHLKBK21}:}
The ZInD dataset features multiple panoramas taken in one room, which could be an excellent resource for self-supervised layout estimation based on multiview stereo. Nevertheless, we do not adopt all panoramas for our pretraining and fine-tuning. The reason is that some of the panoramas include too many occluded corners and current techniques often fail to estimate the layout accurately under such circumstances. ZInD has two attributes that can be used to characterize the complexity of panoramas:
\begin{enumerate}[topsep=0pt]
\item
\emph{Simple}: A panorama is considered simple if it does not contain consecutive occluded corners.
\item
\emph{Primary}: The primary panorama of a room is defined as the best view to annotate, usually the room center.
\end{enumerate}
We filter the data based on these two attributes. First, we exclude scenes that are not `simple'. We then select a pair of panoramas from each room. To ensure better visual overlap for self-supervised training, we require that each pair consists of the primary panorama and a randomly selected non-primary panorama. The target view is randomly assigned at the training stage. We follow the training-test split as ZInD. As a result, we obtain 28{,}828 labeled panoramas for fully-supervised training, 5{,}799 unlabeled pairs of panoramas for self-supervised training, and 2{,}830 labeled panoramas for evaluation.

\paragraph{MatterportLayout~\cite{WangYSCT20,ZouSPCSWCH21}:}
MatterportLayout contains 2{,}295 labeled panorama images capturing real-world Manhattan scenes with different numbers of layout corners.
We subsample the official training split to conduct our data-efficient experiments, and all the results are evaluated on the official test split.

\subsection{Main Results}
\begin{table}[t]
  \centering
  \caption{Improving fine-tuning on ZInD by self-supervised pretraining.}
  \label{tab:ZInD}
\begin{tabular}{c@{\hskip 14pt}c@{\hskip 7pt}|@{\hskip 7pt}c@{\hskip 6pt}c@{\hskip 6pt}c@{\hskip 6pt}c}
\toprule
\makecell{percentage of\\labeled data}  & \makecell{Self-supervised\\pretraining} & 3D IoU(\%)$\uparrow$     & 2D IoU(\%)$\uparrow$   & RMSE $\downarrow$  & $\delta_{1}$ $\uparrow$         \\ \hline
0 & \cmark & 71.59          & 74.57          & 0.3054          & 0.9137          \\ \hline
\multirow{2}{*}{1\% (= 248)} & \xmark      & 69.03          & 72.54          & 0.2165          & 0.8325          \\
                     & \cmark & \textbf{82.33} & \textbf{85.33} & \textbf{0.1847} & \textbf{0.9680} \\ \hline
\multirow{2}{*}{2\% (= 451)} & \xmark      & 82.29          & 85.46          & 0.1945          & 0.9629          \\
                     & \cmark & \textbf{84.17} & \textbf{86.89} & \textbf{0.1673} & \textbf{0.9739} \\ \hline
\multirow{2}{*}{3\% (= 673)} & \xmark      & 84.08          & 86.93          & 0.1733          & 0.9709          \\
                     & \cmark & \textbf{84.65} & \textbf{87.31} & \textbf{0.1648} & \textbf{0.9750} \\ \hline
\multirow{2}{*}{5\% (= 1{,}139)} & \xmark  & 85.28          & 87.20          & 0.1515          & 0.9754          \\
                     & \cmark & \textbf{86.11} & \textbf{88.63} & \textbf{0.1468} & \textbf{0.9791} \\
\bottomrule
\end{tabular}
\end{table}

\paragraph{Improving the fine-tuning on ZInD with self-supervised pretraining.} 
We aim to show that our self-supervised pretraining can provide better weight initialization for fine-tuning the layout estimation model. We divide the experiment into several parts. For each part, we split the training set into labeled data and unlabeled data. ZInD organizes the data in several levels. The top level comprises the scenes, where each scene consists of several rooms. Each room then contains several panoramas sharing the same layout. As our self-supervised pretraining requires pairs of panoramas taken in the same room, we split the training set by the level of scenes, so the self-supervised training pairs adopted in the experiment are guaranteed to be valid.
Therefore, the supervised ratio X\% means that X\% of scenes in the training set are picked as labeled data, while the rest (100-X)\% of scenes are unlabeled data. We use the proposed self-supervised learning mechanism with the unlabeled data to pretrain the model for 20 epochs. As the ZInD is a dataset of unfurnished rooms, leading to faster convergence, we only finetune with the labeled data for 30 epochs. As \cref{tab:ZInD} shows, we set the supervised ratio to 1\%, 2\%, 3\%, and 5\%, corresponding to 248, 451, 673, and 1139 labeled data. We compare our self-supervised model with the ImageNet-pretrained model under each condition. It is clear that our model with self-supervised pretraining outperforms the ImageNet-pretrained model under data-scarce conditions where only a small number of labeled data are available for fine-tuning. Although the performance gap between the two weight initialization strategies decreases as the number of labeled data increases, our self-pretrained model still achieves better performance for all the scenarios.

\begin{table}[t]
  \centering
  \caption{Fine-tuning on MatterportLayout with self-supervised pretraining.}
  \label{tab:Matterport}
\begin{tabular}{c@{\hskip 4pt}c@{\hskip 4pt}c@{\hskip 4pt}c@{\hskip 4pt}c@{\hskip 4pt}c@{\hskip 4pt}c@{\hskip 4pt}c}
\toprule
                           & \multicolumn{5}{c}{3D IoU (\%) $\uparrow$}                                                         \\ \hline
Method                     & 50 labels      & 100 labels     & 200 labels     & 400 labels     & 1650 labels    \\ \hline
ImageNet-pretrained         & 63.28          & 68.15          & 72.35          & 74.36          & 79.70          \\
Self-pretrained & \textbf{65.11} & \textbf{70.18} & \textbf{73.50} & \textbf{76.10} & \textbf{80.90} \\ \hline
                           & \multicolumn{5}{c}{2D IoU (\%) $\uparrow$}                                                         \\ \hline
Method                     & 50 labels      & 100 labels     & 200 labels     & 400 labels     & 1650 labels    \\ \hline
ImageNet-pretrained         & 67.42          & 72.00          & 75.38          & 77.31          & 81.86          \\
Self-pretrained & \textbf{68.65} & \textbf{73.46} & \textbf{76.71} & \textbf{78.97} & \textbf{83.37} \\ \hline
                           & \multicolumn{5}{c}{RMSE $\downarrow$}                                                           \\ \hline
Method                     & 50 labels      & 100 labels     & 200 labels     & 400 labels     & 1650 labels    \\ \hline
ImageNet-pretrained         & 0.527          & 0.446          & 0.371          & 0.344          & 0.280          \\
Self-pretrained & \textbf{0.488} & \textbf{0.411} & \textbf{0.360} & \textbf{0.317} & \textbf{0.256} \\ \hline
                           & \multicolumn{5}{c}{$\delta_{1}$ $\uparrow$}                                                        \\ \hline
Method                     & 50 labels      & 100 labels     & 200 labels     & 400 labels     & 1650 labels    \\ \hline
ImageNet-pretrained         & 0.857          & 0.899          & 0.937          & 0.950          & 0.968          \\
Self-pretrained & \textbf{0.893} & \textbf{0.924} & \textbf{0.947} & \textbf{0.966} & \textbf{0.982} \\ \bottomrule
\end{tabular}
\end{table}

\paragraph{Improving the fine-tuning on MatterportLayout with self-supervision.} 
As aforementioned, the ZInD dataset contains mainly unfurnished rooms. To show that the pretrained model can generalize to furnished rooms, we adopt the self-supervised model that is trained on the entire set of unlabeled data of ZInD to obtain the pretrained weights for the evaluation on MatterportLayout.
We follow the experiment settings of SSLayout360~\cite{Tran21} on MatterportLayout and show the results in \cref{tab:Matterport}. For each experiment, we set the number of labeled images to 50, 100, 200, 400, and 1650 (= the entire labeled data). We select exactly the same labeled subset as SSLayout360. We compare our self-pretrained model with ImageNet-pretrained model as the experiments on ZInD. Again, our method achieves better performance under all conditions and metrics, suggesting that our method can provide effective weight initialization on cross-domain datasets.




\begin{figure}[t]
    \centering
    \begin{tabular}{@{}c@{}}
    {\bf Self-supervised only on ZInD} \\
    \includegraphics[width=.95\linewidth]{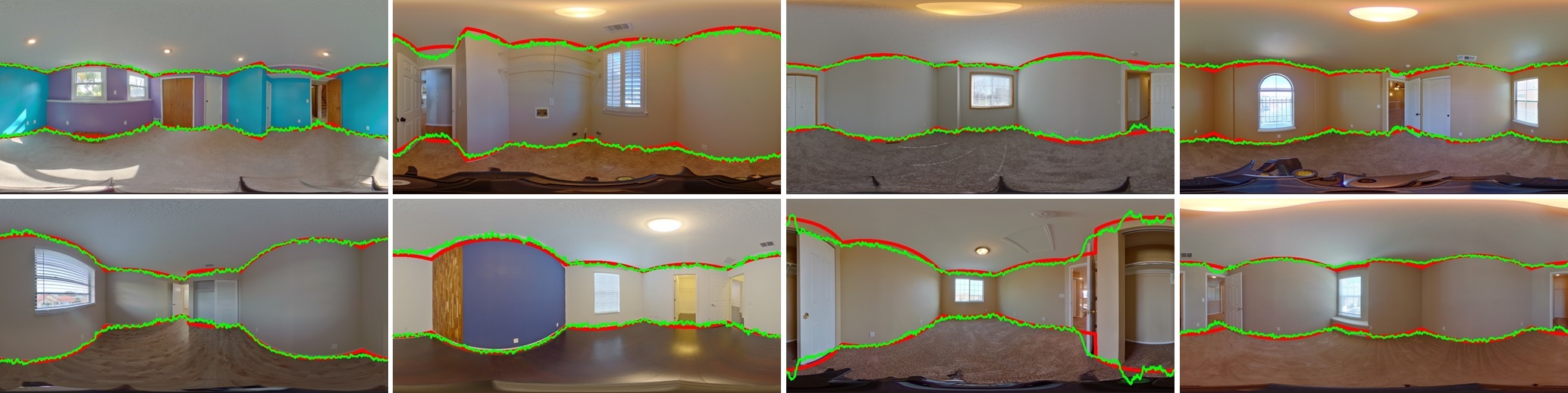} \\
    {\bf Fine-tuning on MatterportLayout with 100 labels} \\
    \includegraphics[width=.95\linewidth]{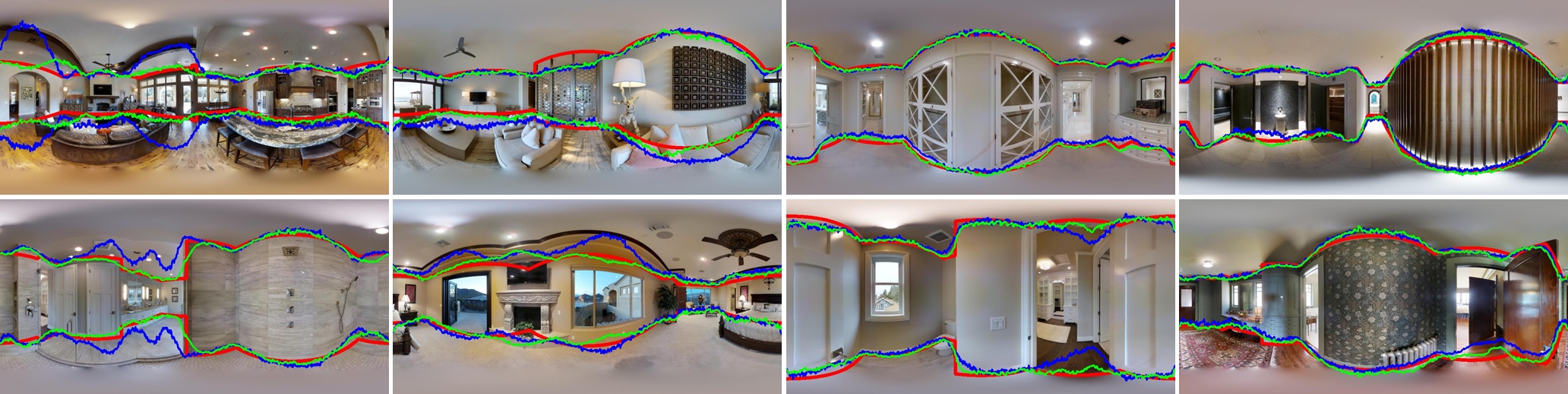} \\
    {\bf Fine-tuning on MatterportLayout with all labels} \\
    \includegraphics[width=.95\linewidth]{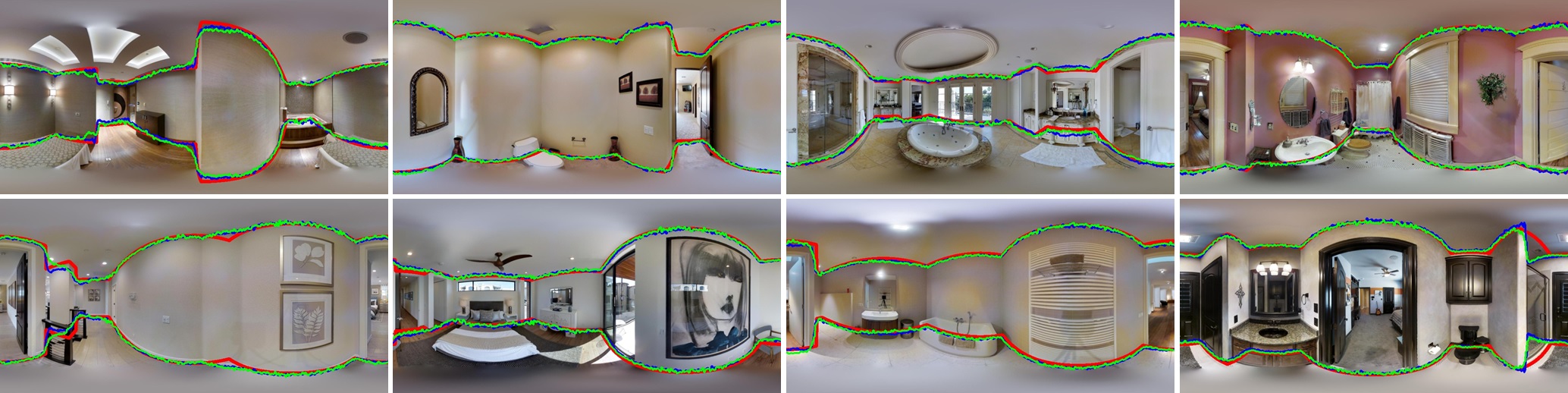}
    \end{tabular}
    \caption{Qualitative results. The \textcolor{red}{red} lines are ground-truth layouts, and the \textcolor{green}{green} lines are our predictions. Our self-supervised model achieves impressive results on the ZInD dataset without training on a single layout label. When fine-tuning on the MatterportLayout dataset, our self-supervised pre-trained achieves better visual results than the ImageNet pre-trained (\textcolor{blue}{blue} line), which is consistent with our quantitative results in \cref{tab:Matterport}.}
    \label{fig:qual}
\end{figure}

\begin{table}[t]
  \centering
  \caption{Active data selection.}
  \label{tab:active_pretrained}
\begin{tabular}{c@{\hskip 12pt}c@{\hskip 12pt}c@{\hskip 6pt}c@{\hskip 6pt}c@{\hskip 6pt}c}
\toprule
&\multicolumn{5}{c}{3D IoU (\%) $\uparrow$}                                                             \\ \hline
Self-pretrained  & Active Selection & 50 labels      & 100 labels     & 200 labels     & 400 labels     \\ \hline
-   & -   & 53.60          & 59.79          & 64.68          & 71.08          \\
-   & \cmark   & \textbf{61.26} & \textbf{65.49} & \textbf{70.14} & \textbf{73.04} \\ \hline
\cmark   & -   & 61.10          & 64.16          & 66.67          & 72.33          \\
\cmark   & \cmark   & \textbf{66.37} & \textbf{69.48} & \textbf{71.44} & \textbf{74.89} \\ \hline
&\multicolumn{5}{c}{2D IoU (\%) $\uparrow$}                                                             \\ \hline
Self-pretrained  & Active Selection & 50 labels      & 100 labels     & 200 labels     & 400 labels     \\ \hline
-   & -   & 58.28          & 64.73          & 69.10          & 74.83          \\
-   & \cmark   & \textbf{65.70} & \textbf{69.68} & \textbf{73.57} & \textbf{76.59} \\ \hline
\cmark   & -   & 64.88          & 67.60          & 70.29          & 75.32          \\
\cmark   & \cmark   & \textbf{69.44} & \textbf{72.88} & \textbf{74.60} & \textbf{77.45} \\ \hline
&\multicolumn{5}{c}{RMSE $\downarrow$}                                                               \\ \hline
Self-pretrained  & Active Selection & 50 labels      & 100 labels     & 200 labels     & 400 labels     \\ \hline
-   & -   & 0.707          & 0.589          & 0.492          & 0.371          \\
-   & \cmark   & \textbf{0.546} & \textbf{0.465} & \textbf{0.392} & \textbf{0.342} \\ \hline
\cmark   & -   & 0.555          & 0.505          & 0.459          & 0.363          \\
\cmark   & \cmark   & \textbf{0.448} & \textbf{0.401} & \textbf{0.375} & \textbf{0.338} \\ \hline
&\multicolumn{5}{c}{$\delta_{1}$ $\uparrow$}                                                            \\ \hline
Self-pretrained  & Active Selection & 50 labels      & 100 labels     & 200 labels     & 400 labels     \\ \hline
-   & -   & 0.794          & 0.821          & 0.875          & 0.942          \\
-   & \cmark   & \textbf{0.834} & \textbf{0.874} & \textbf{0.917} & \textbf{0.950} \\ \hline
\cmark   & -   & 0.868          & 0.894          & 0.903          & 0.950          \\
\cmark   & \cmark   & \textbf{0.910} & \textbf{0.930} & \textbf{0.931} & \textbf{0.960} \\ \bottomrule
\end{tabular}
\end{table}

\paragraph{Active data selection.}
We find that the proposed Manhattan alignment loss and ceiling-floor consistency loss are good indicators for assessing the uncertainties of our self-supervised model's predictions without the need of ground-truth labels. As a result, we can use them to build an efficient scheme for actively selecting the most critical data to be labeled next. Given a pretrained model to be further fine-tuned for layout estimation, we let the model predict on each each sample and calculate the sum of Manhattan alignment loss and ceiling-floor consistency loss as an uncertainty score. If the datasets do not provide annotations of ceiling heights, we estimate the heights using \cref{eq:zc}. For a sample having a higher uncertainty score, we consider its prediction less reliable; it is more useful if we acquire its labeled for the next training phase. To evaluate such an active data selection scheme, we sort the samples by the uncertainty scores in descending order and select the top 50, 100, 200, and 400 samples to form subsets for conducting data-efficiency tests on MatterportLayout. Note that, in addition to the model pretrained by our method, the ImageNet-pretrained model is also evaluated to verify if this data-selection scheme is also applicable to other models.

\cref{tab:active_pretrained} shows that our active data selection scheme is always better than random selection. Furthermore, since the data are selected by the score derived from our self-supervised learning model, \ie, data selection is customized for our self-pretrained model, we find that the fine-tuned model with our data selection scheme outperforms the fine-tuned model with SSLayout360's data split on the first 50 chosen labels as shown in \cref{tab:active_pretrained}. Meanwhile, as SSLayout360 does not mention how their subsequent data splits are chosen, we cannot conduct a fair comparison with SSLayout360's data-scarce results with ours achieved by the active data selection scheme. Nevertheless, it can be seen that our Manhattan alignment loss and ceiling-floor consistency loss are effective to serve as informative metrics for active data selection on layout estimation.

\subsection{Ablation Experiments}
The proposed layout estimation model comprises various learning mechanisms and modules to achieve self-supervision.
Therefore, we conduct extensive ablation experiments to verify the effectiveness of the proposed loss components. We also show the results when room heights and camera heights are not provided.

\begin{table}[t]
\centering
  \centering
  \caption{Ablation of loss components.}
  \label{tab:ablation}
\begin{tabular}{c@{\hskip 4pt}c@{\hskip 4pt}c@{\hskip 4pt}c@{\hskip 4pt}c@{\hskip 4pt}c@{\hskip 4pt}c@{\hskip 8pt}c}
\toprule
photometric & validity & cycle     & src-tgt   & manhattan & ceil-floor & stretch   & 3D IoU(\%)$\uparrow$ \\ \hline
\checkmark   &            &           &           &           &            &           & 53.61 \\ 
\checkmark   & \checkmark  &           &           &           &            &           & 55.43 \\ 
\checkmark   & \checkmark  & \checkmark &           &           &            &           & 56.82 \\ 
\checkmark   & \checkmark  & \checkmark & \checkmark &           &            &           & 64.07 \\ 
\checkmark   & \checkmark  & \checkmark & \checkmark & \checkmark &            &           & 68.68 \\ 
\checkmark   & \checkmark  & \checkmark & \checkmark & \checkmark & \checkmark  &           & 70.21 \\ 
\checkmark   & \checkmark  & \checkmark & \checkmark & \checkmark & \checkmark  & \checkmark & 71.59 \\ \bottomrule
\end{tabular}
\end{table}

\paragraph{Loss components.}
We progressively ablate the proposed losses and present the results in \cref{tab:ablation}.
The photometric loss is the fundamental driving force behind our self-supervised layout estimation model.
By training with only the photometric loss, our model can already learn a rough understanding of room layout to achieve 55.43\% IoU.
We also show that excluding pixels covered by camera tripods is reasonable, and the IoU is degraded without it.
In addition to photometric loss, we use cycle consistency to enrich the supervision coming from the rendered images, which improves our results slightly.



The relative translations between views in the ZInD dataset are typically large; such a wide range of variation in geometry also leads to noticeable view-dependent effects in appearance, \eg, the reflections on the floor vary from place to place.
However, the rendered images via warping are unable to reproduce the view-dependent effect, which may hinder the performance if only the photometric loss is applied.
We thus introduce the geometry-based loss, which encourages the estimated layout polygons from the source and target views to be consistent on the world coordinates.
After adding the source-target consistency loss, our results are improved significantly from $56.82\%$ to $64.07\%$ IoU.
It is worth noting that training with only the source-target consistency loss fails to converge as the layout polygons are sparse and the correspondences between vertices are missing.
As a result, the models still mainly rely on the photometric loss, while the source-target consistency acts as an auxiliary loss.

Finally, the proposed losses further guide the estimated layouts to be {\it i)} Manhattan-world aligned, {\it ii)} consistent between ceiling and floor, and {\it iii)} consistent between different augmented stretches.
Combining the three auxiliary losses, we achieve another significant improvement from $64.07\%$ to $71.59\%$ IoU.

\begin{table}[t]
    \centering
    \caption{The performance of ceiling height inference.}
    \label{tab:guess_ceiling}
\begin{tabular}{c@{\hskip 10pt}c@{\hskip 10pt}c@{\hskip 10pt}c@{\hskip 10pt}c} \\ \toprule
Ceiling height annotation & 3D IoU(\%) $\uparrow$     & 2D IoU(\%) $\uparrow$   & RMSE $\downarrow$  & $\delta_{1}$ $\uparrow$ \\ \hline
\xmark                    & 69.62 & 71.23 & 0.3201 & 0.8988  \\
\cmark                 & \textbf{71.59} & \textbf{74.57} & \textbf{0.3054} & \textbf{0.9137} \\ \bottomrule
\end{tabular}
\end{table}

\paragraph{Inferring the ceiling height.}
The ratio of camera height to ceiling height is important information to reconstruct up-to-a-scale 3D room layout.
\cref{tab:guess_ceiling} presents the results of our self-supervised learning with and without the ceiling height annotation (distance to floor is set to $1$).
Our method can automatically infer the ceiling height if not provided.
The result without ceiling height annotation only degrades slightly and is still ranked the third-best among all results in \cref{tab:ablation}.
Thus, ceiling height annotation is not a crucial requirement for our approach.

\section{Conclusion}
Our method brings self-supervised learning into \threesixty room layout estimation.
We propose a novel \emph{Differentiable Layout View Rendering} module that can warp images from different viewpoints in a differentiable manner, so we can now train \threesixty layout estimation models with image reprojection loss.
We also introduce several {\it a priori} alignment and consistency losses to facilitate this challenging problem and improve the results.
Our self-supervised models are also demonstrated on the practical lower-shot fine-tuning and active-learning setups.
A major limitation of this work is that we only use empty rooms to conduct self-training.
We hope our method can encourage more exploration on self-supervised \threesixty layout estimation and extend to a less constrained data setup.

\clearpage

\section*{Appendices}
The appendices contain \emph{i}) detailed description and illustration of ceiling height inference (\cref{sec:supp_ceil_h}), \emph{ii}) breakdown of quantitative results with different number of corners (\cref{sec:supp_quan_detail}), \emph{iii}) implementation detail for the training loss weights, and \emph{iv}) additional qualitative comparisons for ablation experiments.  

\appendix

\numberwithin{figure}{section}
\numberwithin{table}{section}
\section{Details of ceiling height inference} \label{sec:supp_ceil_h}
\begin{figure}
    \centering
    \begin{tabular}{@{}c@{\hskip 10pt}c@{\hskip 10pt}c@{}}
    \adjincludegraphics[width=.25\linewidth,trim={0 0 {.7\width} 0},clip]{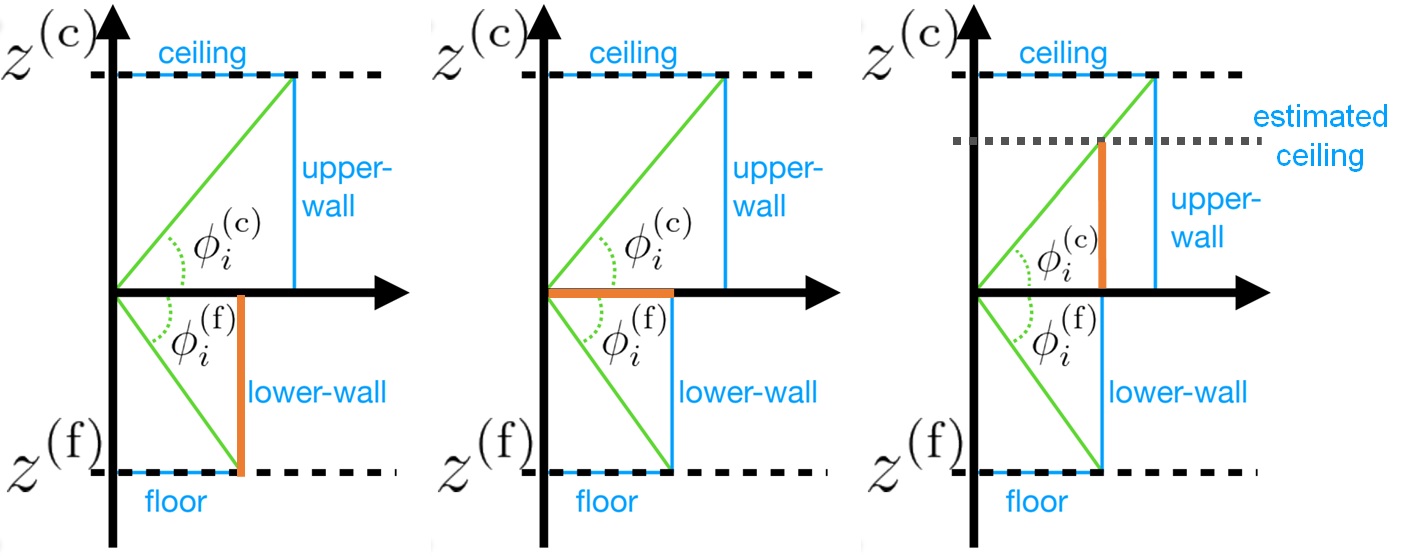} & 
    \adjincludegraphics[width=.25\linewidth,trim={{.3\width} 0 {.4\width} 0},clip]{figures/ceiling.jpg} & 
    \adjincludegraphics[width=.33\linewidth,trim={{.6\width} 0 0 0},clip]{figures/ceiling.jpg} \\
    \textcolor{orange}{$z^{\text{(f)}} = -1$} & 
    \textcolor{orange}{$-\cot{\phi_i^{\text{(f)}}}$} & 
    \textcolor{orange}{$-\cot{\phi_i^{\text{(f)}}}\tan{\phi_i^{\text{(c)}}}$}
    \end{tabular}
    \caption{An illustration of ceiling height inference.}
    \label{fig:ceiling_inference}
\end{figure}


We provide an illustration in \cref{fig:ceiling_inference} to help the reader better understand our ceiling height inference presented in \cref{ssec:DLVR} of the main content.
Recap that we reconstruct up-to-a-scale room layout and set $z^{\text{(f)}} = -1$ (the leftmost plot) in this work.
In case the ratio of camera height to ceiling height is not provided by the dataset, we can infer the distance to ceiling $z^{\text{(c)}}$ using model predictions $\bphi^{\text{(f)}}, \bphi^{\text{(c)}} \in \mathbb{R}^{W}$ for an input image with width $W$.
Specifically, consider only an image column $i$, the optimal distance to ceiling is $z^{\text{(f)}}\cot{\phi_i^{\text{(f)}}}\tan{\phi_i^{\text{(c)}}}=-\cot{\phi_i^{\text{(f)}}}\tan{\phi_i^{\text{(c)}}}$.
We want to find $z^{\text{(c)}}$ that minimizes the mean squared error of the estimator from each column, which is solved by Eq.~(3) in the main paper (provided below for the sake of being self-contained).
\begin{align*}
    z^{\text{(c)}} &= \argmin_{z>0}{ \sum\nolimits_{i=1}^{W}\left(   - \cot{\phi_i^{\text{(f)}}}  \tan{ \phi_i^{\text{(c)}}}  - z \right)^2 } \\
    &=  \frac{1}{W} \sum\nolimits_{i=1}^{W}  -\cot{\phi_i^{\text{(f)}}}   \tan{ \phi_i^{\text{(c)}}}   ~.
\end{align*}

\section{Detailed results for fine-tuning} \label{sec:supp_quan_detail}
\begin{table}[]
    \centering
    \caption{Quantitative results on scenes with different numbers of corners by fine-tuning with 50 labels and 100 labels on MatterportLayout dataset.}
    \label{tab:detail}

\begin{tabular}{ccccccc} \toprule
50 labels           & \multicolumn{6}{c}{3D IoU(\%) $\uparrow$}                                                                          \\ \hline
Method              & 4              & 6              & 8              & 10+            & odd            & overall        \\ \hline
ImageNet-pretrained & 64.04          & 61.32          & 66.46          & 59.80          & 53.77          & 63.28          \\
Self-pretrained     & \textbf{65.65} & \textbf{65.35} & \textbf{68.31} & \textbf{63.24} & \textbf{56.59} & \textbf{65.11} \\ \hline
                    & \multicolumn{6}{c}{2D IoU(\%) $\uparrow$}                                                                          \\ \hline
ImageNet-pretrained & 68.72          & 65.55          & 70.55          & 62.72          & 57.13          & 67.42          \\
Self-pretrained     & \textbf{69.66} & \textbf{68.22} & \textbf{72.65} & \textbf{66.06} & \textbf{59.04} & \textbf{68.65} \\ \hline
                    & \multicolumn{6}{c}{RMSE $\downarrow$}                                                                            \\ \hline
ImageNet-pretrained & 0.506          & 0.487          & 0.547          & 0.626          & 0.606          & 0.527          \\
Self-pretrained     & \textbf{0.463} & \textbf{0.432} & \textbf{0.545} & \textbf{0.563} & \textbf{0.572} & \textbf{0.488} \\ \hline
                    & \multicolumn{6}{c}{$\delta_{1}$ $\uparrow$}                                                                         \\ \hline
ImageNet-pretrained & 0.845          & 0.870          & 0.899          & 0.859          & 0.843          & 0.857          \\
Self-pretrained     & \textbf{0.883} & \textbf{0.901} & \textbf{0.912} & \textbf{0.894} & \textbf{0.911} & \textbf{0.893} \\ \bottomrule
\end{tabular}

\begin{tabular}{ccccccc} \toprule
100 labels          & \multicolumn{6}{c}{3D IoU(\%) $\uparrow$}                                                                          \\ \hline
Method              & 4              & 6              & 8              & 10+            & odd            & overall        \\ \hline
ImageNet-pretrained & 70.78          & 66.63          & 70.81          & 60.48          & 55.09          & 68.15          \\
Self-pretrained     & \textbf{71.26} & \textbf{73.27} & \textbf{71.88} & \textbf{63.96} & \textbf{60.74} & \textbf{70.18} \\ \hline
                    & \multicolumn{6}{c}{2D IoU(\%) $\uparrow$}                                                                          \\ \hline
ImageNet-pretrained & 75.09          & 70.74          & 73.93          & 62.70          & 58.54          & 72.00          \\
Self-pretrained     & \textbf{75.16} & \textbf{75.60} & \textbf{74.78} & \textbf{66.07} & \textbf{63.18} & \textbf{73.46} \\ \hline
                    & \multicolumn{6}{c}{RMSE $\downarrow$}                                                                            \\ \hline
ImageNet-pretrained & 0.394          & 0.408          & 0.521          & 0.604          & 0.578          & 0.446          \\
Self-pretrained     & \textbf{0.380} & \textbf{0.335} & \textbf{0.495} & \textbf{0.518} & \textbf{0.502} & \textbf{0.411} \\ \hline
                    & \multicolumn{6}{c}{$\delta_{1}$ $\uparrow$}                                                                         \\ \hline
ImageNet-pretrained & 0.897          & 0.898          & 0.915          & 0.894          & 0.888          & 0.899          \\
Self-pretrained     & \textbf{0.912} & \textbf{0.933} & \textbf{0.926} & \textbf{0.956} & \textbf{0.956} & \textbf{0.924} \\ \bottomrule
\end{tabular}
\end{table}

We provide the detailed results under label-scarce conditions (fine-tune with 50 labels and 100 labels) shown in \cref{tab:detail}.
We group the results by the ground-truth numbers of corners.
The `overall' column are the results averaged over all samples.
The results show that the self-supervised pretrained models outperform the ImageNet-pretrained model on rooms with different numbers of corners.
Our pre-training can moderately improve results on the simplest cuboid rooms (with only four corners).
On rooms with more complicated topology, our self-supervised pre-training shows significant improvements, especially for rooms with more than ten corners and odd numbers of corners.
The results suggest that the proposed self-supervised approach can serve as a good pre-training strategy and provide a weight initialization that can generalize well under label-scarce conditions.

\section{Detail of training loss weights}
In addition to the main photometric loss $\mathcal{L}_{\text{photo}}$ to guide self-supervised training, we introduce several auxiliary losses to facilitate the challenging self-supervised task.
The training loss weights for the auxiliary losses are set to {0.1}, so the overall training objective is
\begin{equation*}
    \mathcal{L} = \mathcal{L}_{\text{photo}} + 0.1\cdot \left(\mathcal{L}_{\text{cycle}} + \mathcal{L}_{\text{src-tgt}} + \mathcal{L}_{\text{c-f}} + \mathcal{L}_{\text{M}} + \mathcal{L}_{\text{stretch}}\right) \,.
\end{equation*}

\section{Qualitative results of ablation experiments}
\begin{figure}
    \centering
    \includegraphics[width=\linewidth]{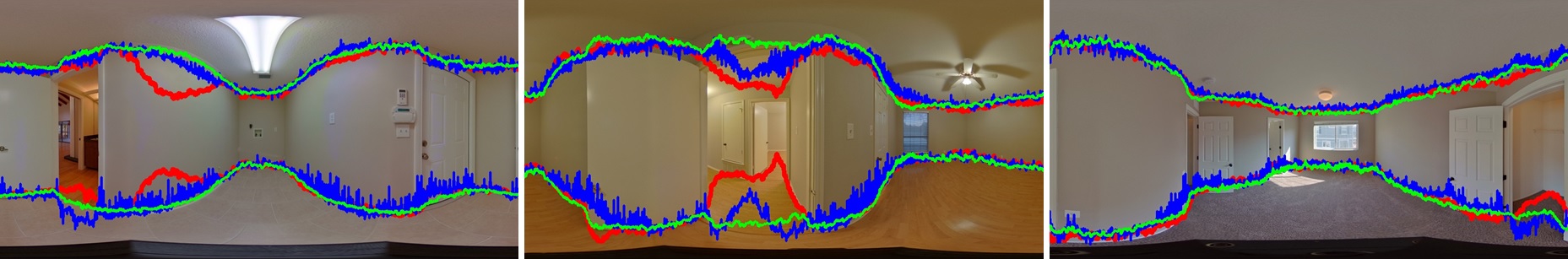}
    \begin{tabular}{@{}c@{\hskip 10pt}c@{\hskip 10pt}c@{}}
    \textcolor{red}{R}: $\mathcal{L}_{\text{photo}}, \mathcal{L}_{\text{cycle}}$. &
    \textcolor{blue}{B}: $\mathcal{L}_{\text{photo}}, \mathcal{L}_{\text{cycle}}, \mathcal{L}_{\text{src-tgt}}$. &
    \textcolor{green}{G}: All losses.
    \end{tabular}
    \caption{Qualitative results of ablation experiments.}
    \label{fig:ablation}
\end{figure}
We provide qualitative comparisons for ablation experiments in \cref{fig:ablation} to further demonstrate the effectiveness of the proposed losses.
The \textcolor{red}{red lines} are the results trained by the photometric loss and the cycle consistency loss, where the model can finally learn a rough understanding of the room layout.
However, there are two main issues when the supervisions come only from the rendered image.
First, the model mistakenly recognizes the space outside doors as part of the room.
Second, image warping cannot reproduce the view-dependent effects in appearances, such as reflection and light conditions.
As a result, the ambiguity of photometric consistency may hinder self-supervised performance.

We thus introduce the source-target consistency to encourage geometric consistency as an auxiliary to the main photometric losses.
We can see from the \textcolor{blue}{blue lines} in \cref{fig:ablation} that the issue of predictions outside doors is substantially addressed.
Consider an example that one of the panoramas is taken in front of an open door, while the other panorama is taken far away from the door. The predicted layout for the first panorama usually has the aforementioned mistake, while the predicted layout for the second panorama may depict the layout along with the door as it is hard to look outside the door from a farther viewpoint.

Finally, the \textcolor{green}{green lines} \cref{fig:ablation} are the results after adding the proposed Manhattan-world alignment, ceiling-floor consistency, and layout stretch consistency losses, which regularize the results to be more in line with the layouts in the real world.
The provided qualitative comparisons explain that each loss in our method is designed not only to improve the quantitative evaluation metrics but also to yield powerful techniques that address the issues in self-supervised training, making our losses useful and explainable.
\clearpage
%
%

\bibliographystyle{splncs04}
\bibliography{egbib}
\end{document}